\pdfoutput=1

\documentclass[a4paper, 10pt, conference]{IEEEtran} 
\usepackage{times}

\usepackage{orcidlink}
\usepackage{graphicx}
\usepackage{epstopdf}
\usepackage{multirow, makecell}
\usepackage{balance}
\usepackage{enumitem}
\usepackage{url}
\usepackage[]{algorithm2e}
\usepackage{amsmath} 

\newcommand\blfootnote[1]{%
  \begingroup
  \renewcommand\thefootnote{}\footnote{#1}%
  \addtocounter{footnote}{-1}%
  \endgroup
}

\renewcommand{\footnoterule}{%
  \kern -3pt
  \hrule width \columnwidth height 1pt
  \kern 2pt
}

\usepackage{color}
\makeatletter
 \let\old@ps@headings\ps@headings
 \let\old@ps@IEEEtitlepagestyle\ps@IEEEtitlepagestyle
 \def\confheader#1{%
 \def\ps@headings{%
 \old@ps@headings%
 \def\@oddhead{\strut\hfill#1\hfill\strut}%
 \def\@evenhead{\strut\hfill#1\hfill\strut}%
 }%
 \def\ps@IEEEtitlepagestyle{%
 \old@ps@IEEEtitlepagestyle%
 \def\@oddhead{\strut\hfill#1\hfill\strut}%
 \def\@evenhead{\strut\hfill#1\hfill\strut}%
 }%
 \ps@headings%
 }
 \makeatother
 
\makeatletter
\def\ps@IEEEtitlepagestyle{
  \def\@oddfoot{\mycopyrightnotice}
  \def\@evenfoot{}
}
\def\mycopyrightnotice{
  {\footnotesize
  \begin{minipage}{\textwidth}
  \centering
  Author's accepted manuscript version. Accepted for publication in the 2021 International Conference on Indoor Positioning and Indoor Navigation (IPIN).
  
  The current author's accepted manuscript version is released for the purpose of meeting public availability requirements. 
   
  Please refer to the published version, for citing this work or for other possible usage. 
  
“\copyright 2021 IEEE.  Personal use of this material is permitted.  Permission from IEEE must be obtained for all other uses, in any current or future media, including reprinting/republishing this material for advertising or promotional purposes, creating new collective works, for resale or redistribution to servers or lists, or reuse of any copyrighted component of this work in other works.”
  
  \end{minipage}
  }
}

\begin{document}


\title{\textbf{ProxyFAUG: Proximity-based Fingerprint Augmentation}\\}

\author{\IEEEauthorblockN{~\large Grigorios G. Anagnostopoulos \orcidlink{0000-0002-8643-7427}, Alexandros Kalousis \orcidlink{0000-0001-6282-0686}}
\IEEEauthorblockA{Geneva School of Business Administration, HES-SO
Geneva, Switzerland\\
Email:  $\left\{ grigorios.anagnostopoulos, alexandros.kalousis \right\}$@hesge.ch}
}

\maketitle

\begin{abstract}
The proliferation of data-demanding machine learning methods has brought to light the necessity for methodologies which can enlarge the size of training datasets, with simple, rule-based methods. In-line with this concept, the fingerprint augmentation scheme proposed in this work  aims to augment fingerprint datasets which are used to train positioning models. The proposed method utilizes fingerprints which are recorded in spacial proximity, in order to perform fingerprint augmentation, creating new fingerprints which combine the features of the original ones. The proposed method of composing the new, augmented fingerprints is inspired by the crossover and mutation operators of genetic algorithms. The ProxyFAUG method aims to improve the achievable positioning accuracy of fingerprint datasets, by introducing a rule-based, stochastic, proximity-based method of fingerprint augmentation. The performance of ProxyFAUG is evaluated in an outdoor Sigfox setting using a public dataset. The best performing published positioning method on this dataset is improved by 40\% in terms of median error and 6\% in terms of mean error, with the use of the augmented dataset. The analysis of the results indicate a systematic and significant performance improvement at the lower error quartiles, as indicated by the impressive improvement of the median error.
\end{abstract}

\renewcommand\IEEEkeywordsname{Keywords}

\IEEEpeerreviewmaketitle

\blfootnote{\\
Code: \url{https://doi.org/10.5281/zenodo.4457353}\\
Data: \url{https://doi.org/10.5281/zenodo.4457391}\\\\
}

%


\section{Introduction} \label{sec:Introduction}


The fingerprint data collection is generally regarded as a tedious and costly process. For this reason, the data collection requirement is considered as a major drawback of fingerprinting methods~\cite{Davidson_Piche_Survey, Hybrid_Generative_Discriminative_2012, Adversarial_Learning, Using_Synthetic_Data, AF-DCGAN}.
The amount of data collected, its geographic distribution and density are factors that play a crucial role in the determination of the achievable performance of positioning systems that build their models by training on these data. Therefore, the ability to meaningfully enrich the volume of the training datasets is a very relevant task.

The recent explosive advancement of the field of Machine Learning (ML), and particularly Deep Learning (DL), is closely related to and facilitated by the vast availability of huge volumes of data. The widely popularised methods of Deep Learning require big data volumes. At the same time, the availability of big data volumes has been one of the determinant enablers of the impressive breakthroughs of DL. Nevertheless, it is often the case that the available datasets over which deep learning models are meant to be trained upon do not offer the variety or the volume that the models would need to perform efficiently. In this context, data augmentation techniques have been widely adopted, in order to provide the data volumes that the models need to reduce their overfitting and to achieve increased accuracy and stability.

Data augmentation techniques aim at increasing the amount of available training data, either by introducing simple rule-based methods that slightly alter existing data, or by synthesizing new data with more elaborate techniques, such as generative modelling. Typical usage of rule-based data augmentation methods are those commonly performed on image datasets~\cite{survey_Image_Augmentation}, where new samples are created by preforming simple actions on existing images, such as: cropping, flipping (horizontally or vertically), rotating, injecting noise, transforming the color distribution, or even mixing images.

Recent works have proposed fingerprinting augmentation methods ~\cite{Permutation_Xiao,Data_Augmentation_Schemes,Sinha_Improved_2020}, evaluating the methods' potential to provide performance improvements, and presenting promising results. Those works rely on the assumption that, at each location appearing on the training dataset, multiple receptions are available. While this assumption is met in some fingerprint datasets, it is not always the case. Such a requirement can be very restrictive, setting a large group of datasets unable to profit from the advantages of fingerprint augmentation.

In this work, the requirement of having multiple receptions recorded at the exact same ground truth location is removed, since ProxyFAUG operates with the notion of spacial proximity, utilizing fingerprints in the same vicinity. The intuition behind the proposed method is rooted on the fact that signal receptions (fingerprints) in locations of close proximity, as well as at the exact same location, might contain unmatching sets of receiving basestations, or have fluctuations on the values of the receiving basestations. Such discrepancies can be expected to appear in various combinations. Exploring a combinatorial approach of altering the existing proximal fingerprints is the path that this work has followed in order to meaningfully augment training datasets.

The rest of this paper is organized as follows. In Section~\ref{sec:Related}, the related work is discussed. Section~\ref{sec:ProxyFAUG} presents ProxyFAUG, the proposed fingerprint augmentation scheme, in its detail. The concise description of the experimental setup, as well as the extensive presentation and discussion of the results is the content of Section~\ref{sec:Results}. Finally, Section~\ref{sec:Conclusions} summarizes this work, presenting conclusions drawn and future directions.

\section{Related Work} \label{sec:Related}

There has been an increasing interest on the subject of fingerprint augmentation over the last years . The existing works on fingerprint augmentation can be categorized into two main approaches. The first one consists of \textit{rule-based fingerprint augmentation} methods, while the second one concerns the generation of synthetic fingerprints through \textit{generative modelling} methods. In this section, we present an overview of the related work of both these approaches.

The methods of augmenting fingerprints through \textit{rule-based} approaches, as the one proposed in the current work, utilize the concepts of adding noise or permuting received values of training fingerprints. Xiao et al.~\cite{Permutation_Xiao} propose a fingerprint augmentation method, exemplifying it using a public Wi-Fi dataset, which has multiple receptions at each training location. Assuming that at each training location a data matrix of size $m$ by $n$ is recorded, corresponding to $m$ receptions for each of the $n$ access points, the authors of~\cite{Permutation_Xiao} propose the generation of a new data matrix of equal size ($m$ by $n$), which is produced upon permuting the elements of each column of the original data matrix. The authors report an improvement, in terms of mean localization error, of approximately 10\%. 

In a recent series of two consecutive works, Sinha et al.~\cite{Data_Augmentation_Schemes,Sinha_Improved_2020}, have proposed three fingerprint augmentation schemes. In the first work~\cite{Data_Augmentation_Schemes}, the authors utilize an indoor dataset in a single floor setting, where a very big volume of data, of over 10000 data points, is collected. Assuming the same $m$ by $n$ data matrix available at each fingerprint, as in~\cite{Permutation_Xiao}, they propose the generation of multiple augmented $m$ by $n$ data matrices, each of which differs from the original matrix simply by the subtraction of a constant number from a random reception reported on the matrix. Alternatively, the authors propose another strategy where a uniformly random value (selected inside the range of values of the certain access point) is used instead of the subtraction of a constant. In their follow-up work, Sinha et al.~\cite{Sinha_Improved_2020} propose a method that offers greater variety of generated fingerprints. In that work, each row of the generated matrices results by randomly selecting one of the $m$ original receptions for each of the $n$ access points, from the real training set. In both these works~\cite{Data_Augmentation_Schemes,Sinha_Improved_2020}, each $m$ by $n$ data matrix creates a grey scale image which feeds a CNN model. Each such image is treated as one fingerprint. The most recent work~\cite{Sinha_Improved_2020} yields the best results, reporting significant improvements in comparison to the case of only using the original training set.

It is noteworthy that the above presented works assume the availability of a multitude of measurements on each of the training locations. Given the fact that the collection of training data is considered a costly and timely process, which is often mentioned as the main barrier of the utilization of fingerprinting techniques, removing such a requirement can be a significant advantage. 
To the best of our knowledge, ProxyFAUG is the first relevant work that does not require multiple measurements per location, and which instead operates with the notion of proximity. Not requiring multiple measurements at the same location in order to perform fingerprint augmentation, allows the application of fingerprint augmentation in a wider range of fingerprint datasets, collected by different surveying techniques. The Sigfox dataset used in this work, along with other Low Power Wide Area Network (LPWAN) datasets which were published together~\cite{Sigfox_Dataset}, were collected with a particular surveying method. More specifically, data were collected by postal service cars which circulated around the city center of Antwerp, while carrying the relevant hardware. Such datasets could profit from proximity based augmentation, as they cannot always meet the strict requirement of multiple measurements per location.

Lastly, the utilization of \textit{generative modelling} in the generation of synthetic fingerprints has recently started getting traction. Zou et al.~\cite{Adversarial_Learning} propose the combination of Conditional Generative Adversarial Networks (GAN) conditioned on the outcome of a Gaussian process regression, to generate fingerprints in indoor locations that were not reachable by a robot which collected the original set of fingerprints. An improvement of 33\% per cent in terms of mean error is reported~\cite{Adversarial_Learning}. Similarly, GAN architectures were also utilized by two other relevant, recent works~\cite{Using_Synthetic_Data,AF-DCGAN}. Nabati et al.~\cite{Using_Synthetic_Data}, exemplify GAN-based fingerprint generation, in a four-room (four-class) classification setting, with 250 training samples for each class. Results indicated that using only 10\% of the training data to synthesize a volume of fingerprints equal to the 90\% of the original dataset, can result in a synthetic dataset that achieves a performance similar to the original dataset.  Li et al.~\cite{AF-DCGAN}, explore GAN-based fingerprint generation, in a limited, 7-by-7 meter setting, exemplifying the feasibility of collection effort reduction without an expense in terms of accuracy.

\section{The Proposed ProxyFAUG Method} \label{sec:ProxyFAUG}

\subsection{The ProxyFAUG Concept} \label{sec:ProxyFAUG_concept}

The main conceptual idea behind the proposed augmentation scheme is presented in this section. It is often the case that fingerprints taken either at exactly the same location or in close proximity present noticeable differences. These differences among two fingerprints may concern either distinct sets of receiving basestations between the two fingerprints, or different measurement values from the same basestation (for instance, in the signal strength, for the RSSI case) among the two fingerprints. Such differences among fingerprints are exemplified in Figure~\ref{fig:crossover}. More particularly, in Figure~\ref{fig:crossover}, Fingerprints $A$ and $B$ have received the same values from the  $1_{st}$ and $3_{rd}$ basestation. Basestation $2$ was heard only from Fingerprint $A$ while basestation $4$ was heard only from Fingerprint $B$ . Basestations $5$ and $6$ were heard in both fingerprints but with different received values. Lastly, the $7_{th}$ and $8_{th}$ basestation were not heard in either of the fingerprints.

Similar fluctuations among proximal fingerprints are a common issue in fingerprint datasets. The idea that ProxyFAUG explores is the creation of new fingerprints, as a combination of signal receptions among proximal fingerprints. The assumption is that a valid fingerprint reception in the area between proximal Fingerprints $A$ and $B$ could contain a combination of values of the two fingerprints. For instance, regarding the example of  Figure~\ref{fig:crossover}, a fingerprint could reasonably have values from both Basestations $2$ and $4$ or from none of them. Such two cases are exemplified by the augmented Fingerprints $Aug_1$ and $Aug_2$ of Figure~\ref{fig:crossover}, which are produced with the ProxyFAUG operators presented bellow.

\begin{table*}[t!]
\begin{center}
    \caption{The Parameters of ProxyFAUG }
    \label{Table:Parameters}
  \begin{tabular}{ | c | p{16cm} | }
    \hline
    \textbf{Parameter} & \multicolumn{1}{|c|}{\textbf{Explanation} } \\ \hline
    $r$& \textbf{Range}: The range (maximum geometric distance) within which  fingerprints will be considered to be proximal \\ \hline
    $S_{max}$ & \textbf{Maximum cluster size}: The maximum number of training fingerprints that will form a cluster, including the fingerprint functioning as a reference from which the  $range$ parameter is measured.\\ \hline
    $N$ & \textbf{Crossovers per pair}: The amount of augmented fingerprints that are produced, by each pair of fingerprints selected from  a cluster. \\ \hline
    $p_m$ & \textbf{Mutation probability}: The probability with which the value of a  cell of the augmented fingerprint gets replaced by a random value. The value is sampled by a uniform distribution in the range between the respective cell values of the two parents of the augmented fingerprint.\\ \hline
  \end{tabular}
\end{center}
\end{table*}

\subsection{The ProxyFAUG Operators} \label{sec:ProxyFAUG_operators}

ProxyFAUG proposes an adjustment of the \textit{crossover} and \textit{mutation} operators of genetic algorithms, in order to implement the rules that will create the augmented fingerprints, in accordance with the logic presented above. In the   \textit{crossover}  operation, two real fingerprints from the training set, \textit{the parents}, are combined to produce a new, augmented fingerprint. ProxyFAUG uses a \textit{uniform crossover}, under which each feature (meaning the reception value of a specific basestation) is chosen from either of the two parents with an equal probability. Traditionally, the crossover operator of genetic algorithms  (often referenced as \textit{recombination} operator as well) implies that each pair of parents generates two complementary offsprings.  ProxyFAUG doesn't impose this restriction and generates one augmented fingerprint per pair of parents, in order to better control the volume of the augmented dataset. Nevertheless, generating two complementary fingerprints for each pair of parents would be a valid option as well.

\begin{figure}[!h]
  \centering
    \includegraphics[width=0.9\linewidth]{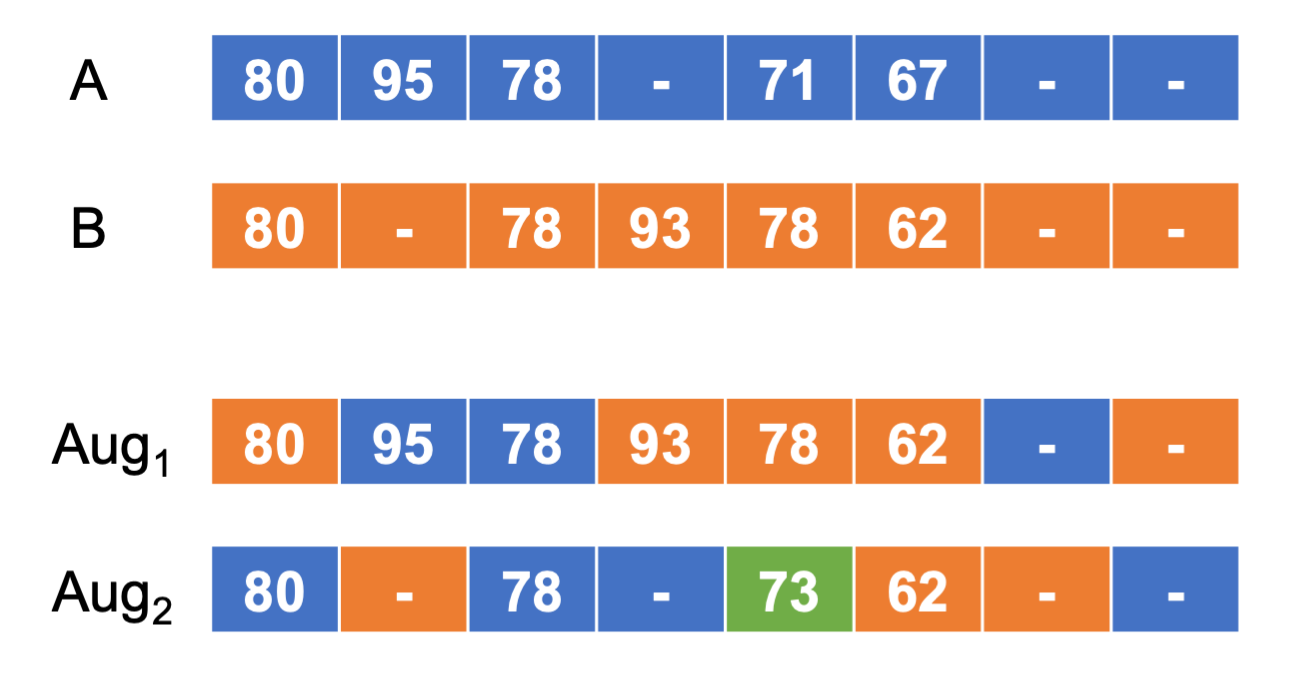}
  \caption{ Example of parent fingerprints A and B producing two augmented fingerprints, by the ProxyFAUG crossover and mutate operators. The augmented fingerprints indicate with the color of each cell, the origin of its value from the respective parent. A cell that underwent a mutation is indicated with green color. }
  \label{fig:crossover}
\end{figure}

The \textit{mutation} operator, in the traditional genetic algorithm setting, concerns the arbitrary flip of a bit in a bit sequence, based on a designer-defined  \textit{mutation probability}.  In ProxyFAUG, we use the \textit{mutation}  operator, in conjunction with the creation of a new, augmented fingerprint by the crossover operator. More specifically, in the augmented fingerprint resulting from the crossover operation, each feature has a probability equal to the  \textit{mutation probability} to be altered. If a feature is mutated, it is replaced by a random value uniformly selected in the range between the respective values of this feature for the two parents. For instance, in Figure~\ref{fig:crossover}, the value of the fifth feature of $Aug_2$ has been mutated. The resulting value $73$ is obtained by a random sampling in the range $[71,78]$, which is defined by the respective values of the fifth feature of the \textit{parent} Fingerprints $A$ and $B$.
The reason for limiting the mutation outcome in this range, is for getting random mutations which range in plausible and meaningful values.

Since the adjusted versions of the two operators are used in conjunction and sequentially in ProxyFAUG, we will refer to their combined usage as \textit{crossover-and-mutate}. Figure~\ref{fig:crossover} exemplifies the two operators, with the creation of two, independent between them, augmented fingerprints. 

The above defined operators describe the part of the augmented fingerprint that concerns the features, meaning the signals that characterize a certain location. Since in ProxyFAUG the fingerprints that are recombined in order to produce new augmented ones have not been necessarily collected at the same location, but in proximity, the location with which the augmented fingerprint is associated with remains to be defined. ProxyFAUG uses the midpoint location of the two parrent fingerprints, to characterize the resulting augmented fingerprint.

\subsection{The ProxyFAUG Proccess} \label{sec:ProxyFAUG_proccess}

In this section, we will present the process that ProxyFAUG follows in order to augment a training set of fingerprints. The process is presented in pseudocode, in Algorithm~\ref{alg:proccess}. The process contains a number of parameters that determine it, and which the designer is called to tune. The parameters, which are summarized in Table~\ref{Table:Parameters}, are subsequently explained in detail.

\begin{algorithm}
\label{alg:proccess}
 \KwIn{$training\_set, r, N, S_{max}, p_m$  }
 \KwOut{$augmented\_training\_set$ }
 \For{$training\_point \in training\_set$}{
  Define $cluster$ of training points in range $r$  from $training\_point$\;
  \If{$cluster$ size  $> S_{max}$}{randomly keep $S_{max}$ points in the $cluster$\; }      
  Define $cluster\_pairs$ containing all pairs of the $cluster$ elements\;  
 \For{$pair \in cluster\_pairs$}{
   Create $N$ augmented fingerprints by the \textit{crossover-and-mutate} operator with $p_m$ on the $pair$\;  
   Add the $N$ augmented fingerprints in the $augmented\_training\_set$\;  
  }
 }
 Copy $training\_set$ in $augmented\_training\_set$\;
 Return $augmented\_training\_set$\;
 \
 \caption{Pseudocode of the ProxyFAUG proccess.}
\end{algorithm}

Initially, the notion of proximity should be concretely defined by the range parameter $r$, which represents the maximum geographic distance between two fingerprints for them to be considered as proximal.
Each fingerprint in the training set will form, together with a set of randomly selected proximal fingerprints, a cluster. 
Since there might be a big amount of fingerprints in the proximity of a certain fingerprint, we define $S_{max}$ as the maximum
cluster size. With this limitation, each training fingerprint will form a cluster together with at most $S_{max}-1$ other fingerprints that exist in a range of $r$ meters from it.
If there are up to $S_{max}-1$ fingerprints in the range of a given fingerprint, they will compose, together with the said fingerprint, a cluster of size that is at most equal to $S_{max}$. Alternatively, if the are $S_{max}$ or more fingerprints in range, a random selection takes place for forming the cluster of size $S_{max}$.

Once the cluster has been defined, all possible pairs of fingerprints are identified. Each pair will function as the \textit{parents} over which the combined \textit{crossover-and-mutation} operator, presented in~\ref{sec:ProxyFAUG_operators}, will be performed. Since a pair might have multiple recombinations that can be useful, a parameter $N$ is available to be tuned by the designer, which will determine the number of augmented fingerprints produced by a pair of \textit{parents}, through the \textit{crossover-and-mutate }operator. As mentioned in Section~\ref{sec:ProxyFAUG_operators}, the midpoint location of the two \textit{parent} fingerprints, is assigned as the ground truth location of each augmented fingerprint.

The process can be summarized as follows. Each fingerprint of the training set composes a cluster of a maximum size $S_{max}$, together with other randomly selected fingerprints of the training set, which should all lay in its vicinity, in a geometric distance that is at most equal to the range parameter $r$. Each possible pair of fingerprints of a cluster produces an amount of augmented fingerprints equal to the parameter $N$, through the \textit{crossover-and-mutate} operator.  

The volume of the resulting augmented dataset, as well as its quality, will depend on the selected parameter values. We will now calculate the maximum size that the augmented training set may have, to obtain an intuition about the impact of the selected parameter values on the resulting dataset size. Let $M$ be the size of the training set. Assume that the range parameter $r$ and the dataset's spatial distribution is such that there exist $S_{max}$ proximal fingerprints for each of the training fingerprints. Then,  $M$ clusters of size $S_{max}$ will be formed. Each cluster will create an amount of pairs of \textit{parents} equal to the number of combinations of $S_{max}$ objects taken 2 at a time. Moreover, each of these pairs will create $N$ augmented fingerprints. Overall, the size of the augmented training set $M_{Aug}$, is upper bounded by the term described in Inequality~\ref{Eq:size}.

\begin{equation}
\label{Eq:size}
M_{Aug}  \leq M + M  {S_{max} \choose 2}  N = M +  M \frac{S_{max}!}{2!(S_{max}-2)!}  N 
\end{equation}

From the above formula, it can be concluded that parameter $N$ shows a linear relation with the resulting volume of the augmented training set. On the contrary, the maximum cluster size parameter $S_{max}$ may affect the size of the augmented training set with a factorial term, and therefore, should be used with caution. 

\section{Experimentation and Results} \label{sec:Results}

\subsection{Setting Selection and Motivation} \label{sec:motivation}

In this section, we will investigate the effect of the ProxyFAUG data augmentation scheme on the performance of a positioning system, utilizing a public dataset of a real-world Sigfox deployment~\cite{Sigfox_Dataset}. The Sigfox dataset used in this study has been collected in the urban area of Antwerp, in Belgium. Datasets of LPWANs, such as this Sigfox dataset, constitute appropriate test-beds for the ProxyFAUG augmentation scheme for following reasons.

Firstly, each recorded fingerprint contains the RSSI values of the transmitted signal as received by each of the 84 basestations that are present in the dataset, plus the spatial ground truth of the signal’s transmission location, as estimated by a GPS device. The fact that an estimate, which is subject to error, is used as ground truth introduces a certain bias. Since the typical localization error of GPS is of a lower order than that of the Sigfox localization, its selection as a ground truth is acceptable~\cite{Sigfox_Dataset}. Having clarified this fact, we proceed with our assumption. The uncertainty of the actual location of the recorded fingerprints facilitates us to make the assumption that, even if fingerprints are not assigned to the exact same locations but in close proximity (based on their GPS-defined ground truth) they could be characterized by similar signal receptions.

Secondly, such big urban datasets are often characterized by an uneven density of recorded fingerprints. Spatial zones with low density of fingerprints might not have enough variance in the training set so as to correctly match a new signal reception with one of the fingerprints of the zone. 
Moreover, since a big amount of basestations is present, it is often the case that distinct sets of basestations are present in fingerprints collected at the same (or in proximal) locations. One could assume that the numerous permutations of the sets of basesations that are present in these fingerprints could constitute valid potential fingerprints, as described in Section\ref{sec:ProxyFAUG_concept} and exemplified in Figure~\ref{fig:crossover}.
The possibility to augment fingerprints in the way proposed by ProxyFAUG can increase the variance of the dataset and enhance the accuracy of a positioning system. Conceptually, ProxyFAUG aims to fill in the gaps of the data collection, an action which seems appropriate in this kind of datasets. 

Lastly, the big volume of data of such datasets, comes hand in glove with the intention of ProxyFAUG to enrich the training set with multiple valid permutations of recorded fingerprints, so as to create a dense mesh of points in the signal space, that would allow new receptions to be correctly affiliated with fingerprints that are close to the actual location under question.
Having presented the intuition behind the setting selection of the mentioned Sigfox dataset for exemplifying and testing ProxyFAUD, we proceed to the experimentation part.

\subsection{Tuning} \label{sec:Tuning}

In order to evaluate the influence of the ProxyFAUG augmentation scheme on a positioning system, we will compare the performance of the same fingerprinting method in two cases: in the case of only using the original training set, against the case where the augmented training set is used. As the positioning method, we will use the one that, to the best of our knowledge, is the best performing published method dealing with the given dataset, presented by Anagnostopoulos and Kalousis~\cite{Anagnostopoulos2019_IPIN}.  As suggested in~\cite{Anagnostopoulos2019_IPIN}, the dataset is initially preprocessed by the $powed$ transformation, which has been initially introduced by Torres-Sospedra et al.~\cite{TORRES_SOSPEDRA_2015}, with the parameter $\beta$ of the $powed$ transform set to 2.6. Moreover, the arbitrary value $-200$ which was set in the original dataset in replacement of the out-of-range missing values is replaced by the experimental minimum received RSSI value of the training set, that is $-157$. Lastly, the Bray-Curtis dissimilarity is used as a distance metric in a kNN setting, were $k=6$ is set. It is worth noting that the same train / validation / test set split used in~\cite{Anagnostopoulos2019_IPIN}, which is publicly available~\cite{Anagnostopoulos2019_IPIN_data}, has been reused in the current work.

\begin{table}[h]
\begin{center}
    \caption{The ProxyFAUG Parameters Selected for this Setting}
    \label{Table:Parameters_selected}
  \begin{tabular}{ | c  | c | }
    \hline
    \textbf{Parameter} & \textbf{Explanation}  \\ \hline
    $r$ & 20 meters \\ \hline
    $S_{max}$ & 2\\ \hline
    $N$ & 8 \\ \hline
    $p_m$ & 0.3\\ \hline
  \end{tabular}
\end{center}
\end{table}

Upon experimentation with the ProxyFAUG parameters, the values reported in Table~\ref{Table:Parameters_selected} were selected for the given setting. The selected parameter values suggest that  each training fingerprint is coupled with  another training fingerprint that is in a range of 20 meters (if such a fingerprint exists). Each such couple of fingerprints, is considered a cluster of $S_{max}=2$. The two fingerprints of the cluster, are the only pair of \textit{parents} of the cluster. Each pair creates $N=8$ augmented fingerprints, by the \textit{crossover-and-mutate} operator, with a mutation probability of $p_m=0.3$. After the augmentation process, the augmented training set is composed of 68127 training points, as opposed to the 10063 points of the original training set.

\begin{figure}[!h]
  \centering
    \includegraphics[width=0.98\linewidth]{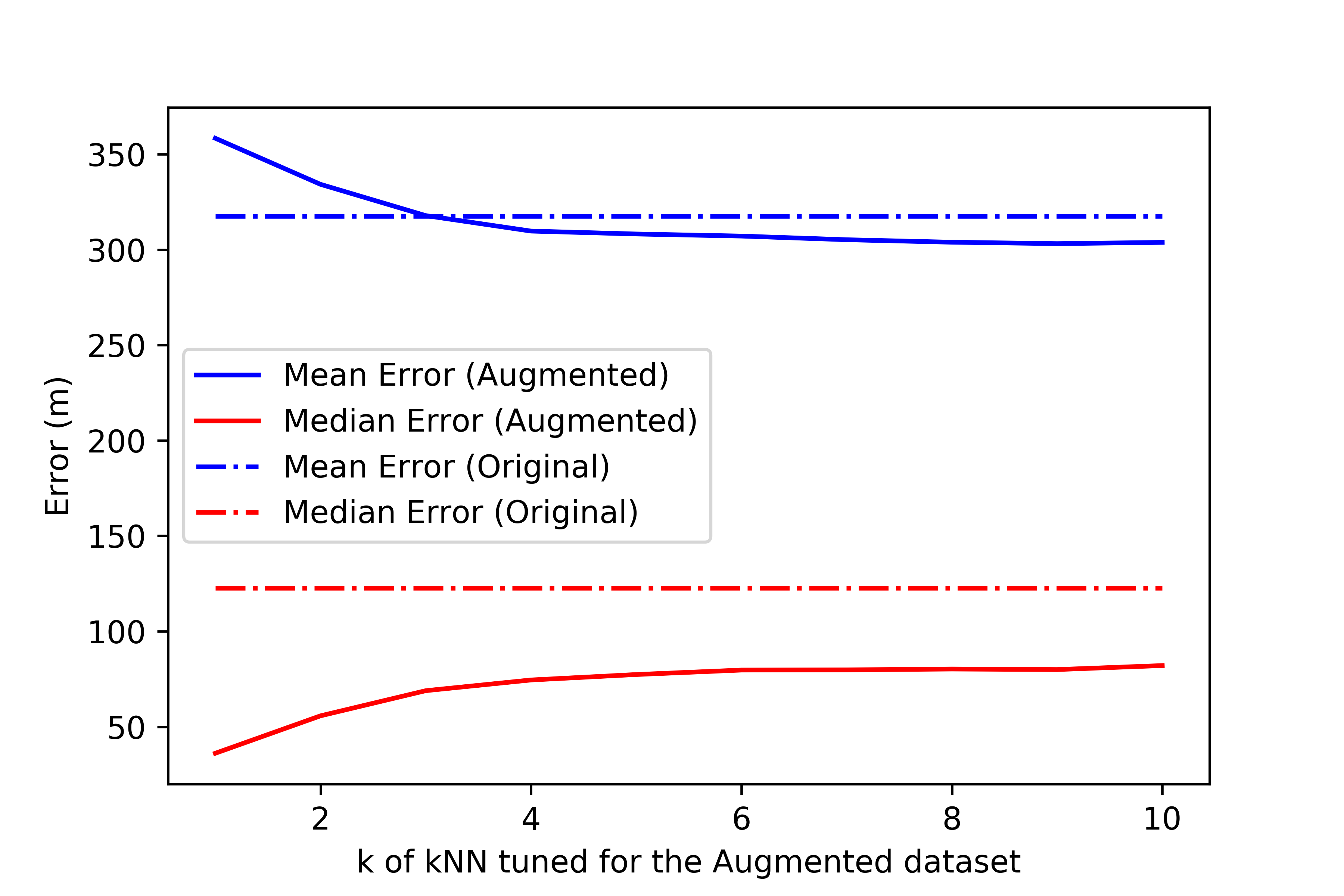}
  \caption{Mean and median localization error on the validation set (in blue and red solid lines respectively) for various value of the k parameter,  for the case of using the augmented training set. The constant value of the best performance on the validation set of the model trained with the original training set is indicated by the dotted lines.}
  \label{fig:tuning}
\end{figure}

Since the landscape of the training set has undegone a significant change with the augmentation process, re-evaluating the hyperparameter selection of the positioning system would be  appropriate. Figure~\ref{fig:tuning} presents the mean and median error on the validation set (with the blue and red continuous line respectively) for various values of $k$, for the case of training on the augmented training set. The dashed lines indicate the level of the performance of the optimal setting of the model trained on the original training set, as proposed in~\cite{Anagnostopoulos2019_IPIN}.

There appears to be a trade-off between the mean and the median error, along the axis representing the amount of $k$ Nearest Neighbours used for the position estimation. The higher values of mean error for low $k$ values can be explained by the fact that, in zones of sparse data presence and for estimates of high localization error, averaging the location of many neighbours may reduce the error. Conversely, the fingerprints of the low quartiles are greatly benefited by the profusion of adequate neighbours, produced by the augmentation process. This fact is highlighted in Figure~\ref{fig:tuning}, which depicts the trend of the median error decreasing as lower values of $k$ are evaluated. The median error goes as down as 36 meters (71\% improvement) for the case of $k=1$, though this setting increases the mean error to 358 meters (13\% deterioration). The value of $k=6$ used for the optimal setting of the original dataset appears to handle well the trade-off, improving both the mean and the median error, and will be used for training on the augmented dataset as well. Therefore, having solidified the hyperparameter setting by evaluating the performance on the validation set, we can proceed to report an unbiased evaluation of the performance on the testing set.

\subsection{Performance} \label{sec:Performance}

\begin{figure}[!h]
  \centering
    \includegraphics[width=0.98\linewidth]{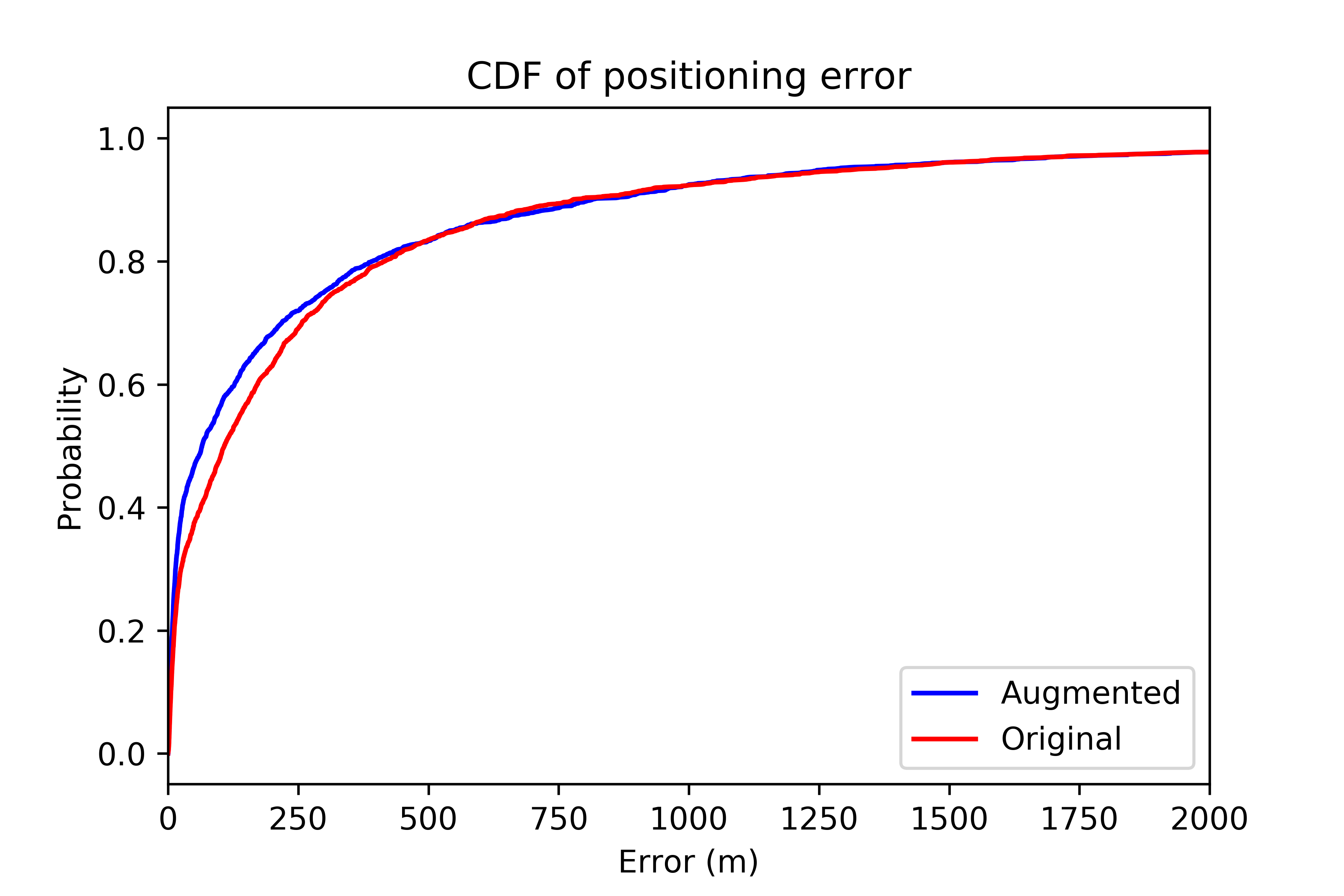}
  \caption{ The CDF of the test error, for the cases of using the original and the augmented training sets respectively.}
  \label{fig:CDF_test}
\end{figure}

The performance of the positioning system for the two cases of using the original and the augmented dataset is presented in Table~\ref{Table:overall_performance} and Figure~\ref{fig:CDF_test}. Table~\ref{Table:overall_performance} reports the mean, median and 75th percentile of error, on the validation and the test set, and for two different training sets: the original and the augmented one. While for the mean and the 75th percentile of error the relative improvement is small, ranging around 6\%, the improvement in terms of median error is impressive. More specifically, the use of the augmented dataset improves the median of the test set error by 40\%.

\begin{table}[h]
\caption{Localization Error Comparison Between the Original and the Augmented Training Set in Terms of Validation and Test Error}\label{Table:overall_performance}
\centering\setlength{\extrarowheight}{2pt}
\centering
\begin{tabular}{|*{7}{c|}}
\hline
 \multirowcell{2}{Training set} &  \multicolumn{3}{c|}{Validation Set Error} & \multicolumn{3}{c|}{Test Set Error} \\
 \cline{2-7}
& \makecell{\textbf{mean}} 
& \makecell{\textbf{median}}
& \makecell{\textbf{75th}} 
& \makecell{\textbf{mean}}
& \makecell{\textbf{median}} 
& \makecell{\textbf{75th}}\\ \hline
\makecell{Original}
& \makecell{318} & \makecell{123} & \makecell{336} & \makecell{298} & \makecell{108} & \makecell{319} \\ 
\makecell{Augmented}
& \makecell{307} & \makecell{80} & \makecell{324} 
& \makecell{280} & \makecell{65} & \makecell{300} \\  \hline
\makecell{\textbf{Improvement}}
& \makecell{\textbf{3\%}} & \makecell{\textbf{35\%}} & \makecell{\textbf{4\%}} &
 \makecell{\textbf{6\%}} & \makecell{\textbf{40\%}} & \makecell{\textbf{6\%}} \\ 
\hline
\end{tabular}
\end{table}
 
The discrepancy of the relative improvement between the mean and median error, is due to the fact that the ProxyFAUG augmentation appears to significantly improve the low quartiles of error in the studied setting. Moreover, the median error, as a metric, is resilient to the presence of outliers of high error that exist when using either of the two training sets. On the contrary, the mean error is mildly affected by the significant improvement of the low ranging errors, as the fewer cases of very high values of error affect it disproportionally. The significant performance improvement of ProxyFAUG in the lower quartiles of error is highlighted in the Cumulative Distribution Function (CDF) of the test error, of Figure~\ref{fig:CDF_test}.

\section{Conclusions and Future Work}
\label{sec:Conclusions}

In this work, we presented ProxyFAUG, a novel, concise methodology of proximity-based fingerprint augmentation. Unlike other common approaches of fingeprint augmentation, ProxyFAUG does not require multiple signal recordings at each training location, a fact that broadens the set of fingerprint datasets in which the augmentation method is applicable. ProxyFAUG proposes a set of parameters, which can allow the system designer to adapt the augmentation method to the particularities of the dataset and to the system requirements in terms of acceptable training data volume,  complexity and achievable localization accuracy. 

The results of the tests of this work, performed in a public, real-world dataset, highlight the significant performance improvements that are achievable with ProxyFAUG. More particularly, ProxyFAUG offers systematic performance improvements in the low quartiles of error, indicated by the 40\% reduction of the median error in the independent test set, in comparison to the best performing published method on the same dataset. The CDF of error, depicted in Figure~\ref{fig:CDF_test}, highlights a clear performance discrepancy between the augmented and the original dataset.

Moreover, a significant difference in the relative improvement between the median and mean error has been observed. A relevant point is the fact that the outliers of high error, that exist when using either the original or the augmented dataset, greatly affect the mean error, while they don't have the same impact on the median metric. In addition, it was observed that ProxyFAUG provides significant and systematic performance improvements in the lower error quartiles, which may be witnessed in the CDF and the median error improvement.

Overall, ProxyFAUG presents a generic parametric method, which allows the designer to appropriately tune its functionality. As a future work, we plan to explore the possibility of alternative, dynamic versions of ProxyFAUG, in which the parameter values used will automatically adjust to the spatial characteristics of the training sets. For instance, the different spatial density of the training sets in different zones, could be used to dynamically adjust the ProxyFAUG parameters values, of parameters such as the range $r$ or the amount of augmentations $N$. The assumption behind this future work direction is that a more targeted and customized arrangement of the augmentation process could potentially offer greater performance improvements. Lastly, we intent to explore the performance improvements of ProxyFAUG with other positioning methods and on other public fingerprint datasets.

It is noteworthy that in the spirit of repeatability, reproducibility, verifiability and comparability of results, the full code implementation of the current work as well as the utilized and resulting datasets are openly available at the Zenodo repository.


\section*{Acknowledgements}
\label{sec:Acknowledgements} 
This work was funded by the Swiss National Science Foundation, under the Spark Funding scheme, in the context of the project Eratosthenes
(project number 195964).

%
\balance

\bibliographystyle{IEEEtran}
\bibliography{bibliography}

\end{document}